\documentclass[runningheads]{llncs}
\usepackage[T1]{fontenc}
\usepackage{graphicx,verbatim}
\usepackage{hyperref}
\usepackage{color}

\begin{document}
\title{RetSTA: An LLM-Based Approach for Standardizing Clinical Fundus Image Reports}
%
% \begin{comment}  %% Removed for anonymized MICCAI 2025 submission
\author{Jiushen Cai\inst{1} \and
Weihang Zhang\inst{1} \and
Hanruo Liu\inst{2,3,1} \and
Ningli Wang\inst{2,4} \and
Huiqi Li\inst{1}}
\authorrunning{Jiushen Cai et al.}
\institute{Beijing Institute of Technology, Beijing, China \and
Beijing Tongren Hospital, Beijing, China \and
Capital Medical University, Beijing, China \and
Beijing Institute of Ophthalmology, Beijing, China \\
\email{zhangweihang@bit.edu.cn}}

% \end{comment}

% \author{Anonymized Authors}  %% Added for anonymized MICCAI 2025 submission
% \authorrunning{Anonymized Author et al.}
% \institute{Anonymized Affiliations \\
%     \email{email@anonymized.com}}

\maketitle              % typeset the header of the contribution
\begin{abstract}
Standardization of clinical reports is crucial for improving the quality of healthcare and facilitating data integration. The lack of unified standards, including format, terminology, and style, is a great challenge in clinical fundus diagnostic reports, which increases the difficulty for large language models (LLMs) to understand the data. To address this, we construct a bilingual standard terminology, containing fundus clinical terms and commonly used descriptions in clinical diagnosis. Then, we establish two models, RetSTA-7B-Zero and RetSTA-7B. RetSTA-7B-Zero, fine-tuned on an augmented dataset simulating clinical scenarios, demonstrates powerful standardization behaviors. However, it encounters a challenge of limitation to cover a wider range of diseases. To further enhance standardization performance, we build RetSTA-7B, which integrates a substantial amount of standardized data generated by RetSTA-7B-Zero along with corresponding English data, covering diverse complex clinical scenarios and achieving report-level standardization for the first time. Experimental results demonstrate that RetSTA-7B outperforms other compared LLMs in bilingual standardization task, which validates its superior performance and generalizability. The checkpoints are available at \url{https://github.com/AB-Story/RetSTA-7B}.

% Authors must provide keywords and are not allowed to remove this Keyword section.

\end{abstract}
\section{Introduction}

\begin{figure}[bt]
\centering
\includegraphics[width=\linewidth]{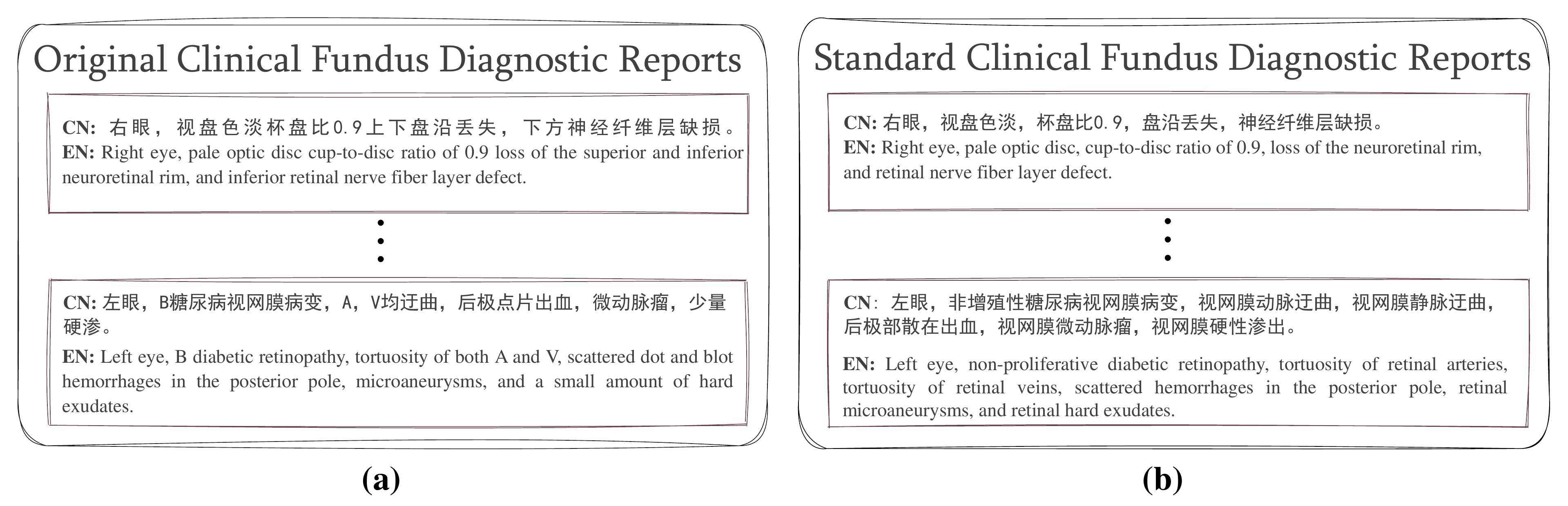}
\caption{The importance of standardization: \textbf{(a)} Non-standardized reports lead to difficulties in comprehension. \textbf{(b)} Standardized reports enhance clarity and accuracy of understanding.} 
\label{fig1}
\end{figure}

Clinical diagnostic reports are crucial records in medical practice, where their accuracy and clarity play a decisive role in patient diagnosis and treatment. Standardized clinical reports can reduce communication costs among healthcare professionals, minimize the potential for misunderstandings, and ensure medical quality and patient safety \cite{shweikh2023growing}. They can be used to generate structured electronic medical records, providing data support for clinical research and labeled data for the development of artificial intelligence-based algorithms. Furthermore, standardized reports can facilitate the integration and analysis of medical data across multiple institutions, providing high-quality input for data mining and LLM development \cite{halfpenny2022towards,hoffmann2024streamlining,domalpally2024data,sheehan2016improving}. Modern ophthalmology research requires large volumes of high-quality data increasingly. It not only expands the available sample size for complete data analysis, but also significantly improves the quality, reliability, and efficiency of ophthalmic research endeavors \cite{wilkinson2016fair}. As shown in Figure \ref{fig1}, the current lack of unified standards in the preparation of clinical diagnostic reports leads to significant variations in format, writing style, and terminology usage, which complicates the understanding of clinical data \cite{hoffmann2024towards,jin2022artificial,cai2023advancing}.

In clinical diagnostic process for fundus conditions, existing ophthalmic clinical guidelines and terminology fail to cover all scenarios encountered. Detailed descriptions of fundus disease symptoms heavily rely on the interpreting experience of ophthalmologists. Furthermore, the prevalence of negative and ambiguous expressions, typos, and issues related to system storage and formatting errors in reports exacerbate the challenges in standardizing fundus diagnostic reports.

Fortunately, the recent emergence of LLMs has introduced new research perspectives to this issue \cite{xie2024preliminary,zhong2024evaluation}. LLMs such as ChatGPT \cite{brown2020language} and LLaMA \cite{touvron2023llama} have demonstrated significant potential across various domains, leveraging their robust capabilities in natural language understanding and generation. DeepSeek has successively released multiple models, including DeepSeek-v3 \cite{liu2024deepseek}, which rivals the performance of GPT-4o \cite{achiam2023gpt} at an exceptionally low cost. In medical field, LLMs like UltraMedical \cite{zhang2024ultramedical} and OpenBioLLM \cite{pal2024openbiollms} have been utilized for tasks such as diagnosis, clinical report generation, and medical Q\&A. LLMs have also demonstrated significant potential in the field of ophthalmology, offering promising prospects for advancing both ophthalmic research and clinical practice \cite{tan2023artificial,chen2024eyegpt,chen2024icga,chen2024ffa,chen2024chatffa,li2024visionunite,li2024integrated}. These advancements indicate that LLMs hold vast application prospects in the standardization of clinical diagnostic reports, addressing challenges arising from inconsistent report formats and terminology effectively.

To take the advantages of LLMs, we propose a novel approach to standardize fundus diagnostic reports. Our main contributions are as follows: 
Firstly, we construct a bilingual standard terminology for fundus clinical practice. This terminology integrates diverse clinical guidelines, standardized terms, and real diagnostic reports to ensure comprehensiveness and practicality. %%%
Secondly, we propose RetSTA-7B-Zero, a LLM specifically designed to comprehend complex expressions in fundus clinical diagnostic scenarios. This model demonstrates exceptional performance in handling intricate standardization tasks.  %%%
Lastly, we propose RetSTA-7B, which is fine-tuned on 60,037 bilingual 'original report-standard report' pairs, ensuring its domain adaptability and disease coverage. To the best of our knowledge, this represents the first attempt to develop a standardization model at the report level, not even limited to ophthalmology. %%

\section{Method}
\subsection{Standard Terminology}
\begin{figure}[bt]
\centering
\includegraphics[width=10.2cm]{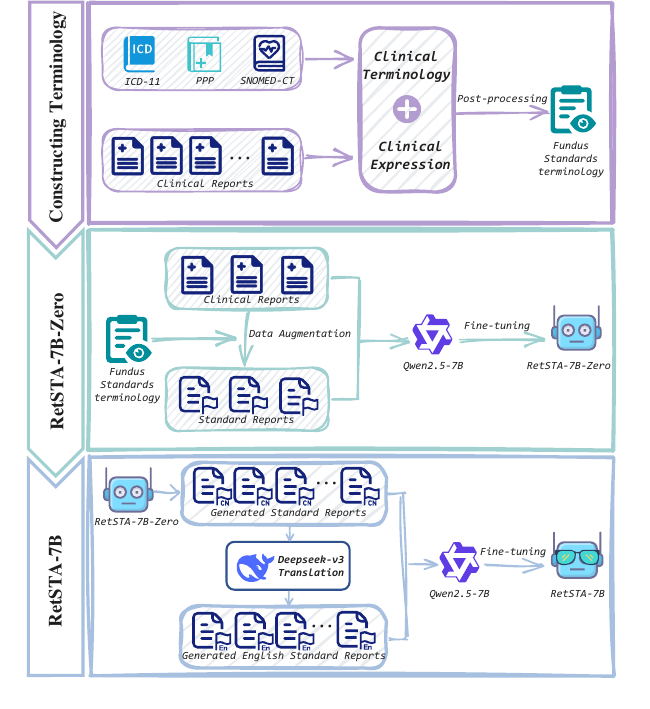}
\caption{Overview of the diagnostic reports standardization paradigm} 
\label{fig2}
\end{figure}

As shown in Figure \ref{fig2}, the fundus diagnostic terminology constructed in this paper comprehensively references international authoritative medical standards and real clinical diagnostic reports. Based on the International Classification of Diseases – 11 (ICD-11), Preferred Practice Pattern (PPP) issued by the American Academy of Ophthalmology, and the Systematized Nomenclature of Medicine – Clinical Terms (SNOMED-CT), the core terminology for fundus diseases was systematically integrated, encompassing fundamental aspects such as disease names, clinical symptoms, and critical examination indicators. By referencing these authoritative standards, the terminology ensures normative and international compatibility. Given that descriptions of certain symptoms in clinical practice are characterized by detail and dynamism, which standard terminology often fails to capture fully, we extracted frequently occurring descriptive phrases from 451,956 real fundus diagnostic reports. These terms were preliminarily screened for accuracy, clarity, and clinical relevance, serving as a critical supplement to the standardized terminology.

To ensure the practicality and reliability of the terminology, we conducted rule-based matching and manual review of candidate terms under the guidance of a senior ophthalmologist. This process aims to eliminate duplicates or ambiguous expressions, and potentially misleading terms, preventing the coexistence of multiple terms for the same clinical entity and retaining only those with clear semantics. As a result, we successfully constructed a bilingual fundus standard clinical terminology, comprising 362 standardized terms.

\subsection{Construction of RetSTA-7B-Zero}
Fine-tuning LLMs for standardizing fundus clinical reports requires a substantial amount of labeled data, while in practical research, acquiring large-scale task-relevant labeled data presents significant challenges. To alleviate this issue, we constructed 4,000 pairs of 'original report-standard report' pairs, with 2,000 pairs serving as an independent test set to evaluate the model's generalization performance, and the remaining 2,000 pairs forming the foundational training set. 

To further expand the data scale, we designed a simple yet effective data augmentation strategy that accounts for various forms of noise prevalent in clinical scenarios, including typos, missing punctuation, ambiguous expressions, and negations. Building upon the foundational training set, this strategy employed techniques such as syntactic noise injection, semantic perturbation, and synonym replacement to expand the data volume to 10,000. Syntactic noise injection simulates real typos by inserting typos randomly, deleting punctuation, or splitting long sentences within the text. Semantic perturbation enhances the model's ability to understand complex semantics by adding negation words or probabilistic modifiers. Synonym replacement utilizes a thesaurus of similar words to substitute disease descriptions, further increasing data diversity. We used the Low-Rank Adaptation (LoRA) \cite{hu2021lora} to fine-tune the Qwen2.5-7B-Instruct \cite{yang2024qwen2} model on the augmented dataset, resulting in the preliminary standardized model, RetSTA-7B-Zero.

\subsection{RetSTA-7B for Standardization}
To further enhance the model's coverage of clinical long-tail scenarios, we utilized RetSTA-7B-Zero to perform batch inference on 60,000 original fundus diagnostic reports, generating candidate standardized reports. To further improve the quality of the generated reports, we filtered out entries containing obvious logical conflicts or terminology errors through rule-based filtering first. Then, we calculated text similarity and removed redundant data with description homogeneity above a threshold.

To enhance the model's cross-lingual adaptability and further improve its capability in medical scenarios, DeepSeek-V3 was employed to translate the Chinese dataset into an English version. During the translation process, we additionally implemented terminology consistency checks to ensure the translation accuracy, thereby mitigating semantic distortion caused by linguistic differences effectively. Ultimately, a mixed dataset containing 60,037 pairs of bilingual reports was constructed. This dataset not only covers standardized descriptions of various fundus diseases but also possesses cross-lingual compatibility. Based on the above dataset, we fine-tuned the Qwen2.5-7B-Instruct model using LoRA, ultimately developing RetSTA-7B, a LLM specifically tailored for the standardization of clinical fundus diagnostic reports.% Through its fine-tuning process, RetSTA-7B has learned the generation logic of standardized reports, demonstrating its capability to effectively process complex expressions encountered in real clinical scenarios and produce standardized diagnostic reports. The implementation of a multi-phase data construction and cleaning strategy ensured the high quality and diversity of the training data, providing a solid foundation for the model's performance enhancement.

\section{Experiment}
\subsection{Experimental Setup}
\subsubsection{Datasets} Two datasets were used for model fine-tuning in the experiments: a dataset of 10,000 samples for fine-tuning RetSTA-7B-Zero, and a bilingual dataset of 60,037 samples for fine-tuning RetSTA-7B. Additionally, an independent test set of 2,000 samples was used to evaluate model performance. For the evaluation of other LLMs, a unified 5-shot learning strategy was adopted, with the standard terminology provided as a reference during testing to ensure consistency and fairness.
\subsubsection{Evaluation Metrics} To comprehensively evaluate the performance of the mod-
el in the task of standardizing clinical fundus diagnostic reports, this paper employed commonly used metrics in the field of Natural Language Generation (NLG), including BLEU-1, BLEU-4, ROUGE-L, and METEOR.
\subsubsection{Implementation Details}
We selected the 7B version of the pre-trained Qwen-
2.5 model, obtaining its checkpoints directly from the official Huggingface repository. All models were trained on NVIDIA Geforce RTX 4090 GPU. The RetSTA-7B-Zero model was fine-tuned for 1 epoch, while the RetSTA-7B model was fine-tuned for 3 epochs. During fine-tuning, the batch size was set to 2, and the gradient accumulation steps were set to 8. We used the AdamW optimizer with 0.03 warm-up ratio and a learning rate of 3e-5. 

\begin{table}[ht]
\centering
\setlength{\tabcolsep}{6pt}
\renewcommand{\arraystretch}{1.2}
\caption{Results on English clinical diagnostic reports.}
\label{tab1}
\begin{tabular}{ccccc}
\hline
& \textbf{BLEU-1} & \textbf{BLEU-4} & \textbf{METEOR} & \textbf{ROUGE-L} \\ \hline
\multicolumn{5}{c}{\textit{\textless 10B Large Language Models}}                               \\
UltraMedical-8B \cite{zhang2024ultramedical}       & 26.20           & 22.01           & 35.33           & 25.09            \\
OpenBioLLM-8B \cite{pal2024openbiollms}         & 61.82           & 47.17           & 52.01           & 44.02            \\
Yi-1.5-9B-Chat \cite{young2024yi}       & 70.47           & 56.86           & 59.29           & 52.58            \\
GLM-4-9B \cite{glm2024chatglm}             & 76.40           & 62.18           & 59.76           & 57.46            \\
Baichuan2-7B-Chat \cite{yang2023baichuan}    & 75.49           & 59.79           & 60.57           & 52.55            \\
Llama-3.1-8B-Instruct \cite{dubey2024llama} & 53.01           & 43.29           & 53.27           & 43.85            \\
Qwen2.5-7B-Instruct \cite{yang2024qwen2}  & 20.23           & 16.10           & 18.77           & 16.03            \\
\textbf{RetSTA-7B}    & \textbf{92.69}  & \textbf{89.66}  & \textbf{86.04}  & \textbf{90.08}   \\ \hline
\multicolumn{5}{c}{\textit{\textgreater 10B Large Language Models}}                            \\
Qwen2.5-72B-Instruct \cite{yang2024qwen2} & 80.32           & 66.65           & 66.69           & 57.27            \\
GLM-4-Plus \cite{glm2024chatglm}           & 70.44           & 54.53           & 53.46           & 52.52            \\ 
DeepSeek-V3 \cite{liu2024deepseek}          & 84.39           & 76.12           & 73.69           & 74.55            \\ 
\hline
\end{tabular}
\end{table}

\subsubsection{Baselines}
We compared our models with two types of LLMs: 1) Small LLMs: UltraMedical-8B \cite{zhang2024ultramedical}, OpenBioLLM-8B \cite{pal2024openbiollms}, Yi-1.5-9B \cite{young2024yi}, GLM-4-9B \cite{glm2024chatglm}, LLaMA-3.1-8B \cite{dubey2024llama}, Qwen-2.5-7B-Instruct \cite{yang2024qwen2}, Baichuan2-7B \cite{yang2023baichuan}; and 2) Large LLMs: Qwen2.5-72B \cite{yang2024qwen2}, GLM-4-Plus \cite{glm2024chatglm} and DeepSeek-V3 \cite{liu2024deepseek}, where Ultra-
Medical-8B \cite{zhang2024ultramedical} and OpenBioLLM-8B \cite{pal2024openbiollms} are medical-specific LLMs.

\subsection{Experimental Results}
RetSTA-7B significantly outperforms the other nine compared LLMs on the test set, including two medical-specific LLMs and three LLMs with large-scale, highlighting its robust capability in the standardization task of clinical diagnostic report.

\begin{table}[bt]
\centering
\setlength{\tabcolsep}{5pt}
\renewcommand{\arraystretch}{1.2}
\caption{Results on Chinese clinical diagnostic reports.}
\label{tab2}
\begin{tabular}{ccccc}
\hline
& \multicolumn{1}{l}{\textbf{BLEU-1}} & \multicolumn{1}{l}{\textbf{BLEU-4}} & \multicolumn{1}{l}{\textbf{METEOR}} & \multicolumn{1}{l}{\textbf{ROUGE-L}} \\ \hline
\multicolumn{5}{c}{\textit{\textless 10B Large Language Models}}                                                                                                                   \\
Yi-1.5-9B-Chat \cite{young2024yi}           & 70.31                               & 56.45                               & 75.76                               & 71.37                                \\
GLM-4-9B \cite{glm2024chatglm}                 & 74.77                               & 60.23                               & 77.60                               & 75.17                                \\
Llama-3.1-8B-Chinese-Chat \cite{dubey2024llama} & 79.02                               & 66.11                               & 80.21                               & 78.88                                \\
Qwen2.5-7B-Instruct \cite{yang2024qwen2}      & 76.83                               & 65.09                               & 79.81                               & 77.99                                \\
Baichuan2-7B-Chat \cite{yang2023baichuan}        & 73.46                               & 57.11                               & 73.94                               & 72.26                                \\
RetSTA-7B-Zero            & 91.93                               & 87.89                               & 92.74                               & 92.77                                \\
\textbf{RetSTA-7B}        & \textbf{93.35}                      & \textbf{89.83}                      & \textbf{94.19}                      & \textbf{94.30}                       \\ \hline
\multicolumn{5}{c}{\textit{\textgreater 10B Large Language Models}}                                                                                                                \\
Qwen2.5-72B-Instruct \cite{yang2024qwen2}     & 81.97                               & 72.39                               & 84.34                               & 83.21                                \\
GLM-4-Plus \cite{glm2024chatglm}               & 83.00                               & 74.21                               & 85.68                               & 84.41                                \\
DeepSeek-V3 \cite{liu2024deepseek}              & 82.92                               & 74.75                               & 83.97                               & 84.35                                \\ \hline
\end{tabular}
\end{table}

\textbf{For English reports}, the results are shown in Table~\ref{tab1}. The results indicate that medical-specific LLMs, like UltraMedical \cite{zhang2024ultramedical} and OpenBioLLM \cite{pal2024openbiollms}, specialize in medical Q\&A but struggle on the standardization task, even when the task is medically related. This suggests that standardization task of clinical diagnostic reports requires not only medical knowledge but also a deep understanding of the inherent logic and structure of text reports. It is noteworthy that despite the implementation of English prompts with explicit language constraints for English responses, Qwen2.5-7B-Instruct \cite{yang2024qwen2} generated outputs containing substantial Chinese content consistently. This linguistic inconsistency significantly compromised the model's performance in standardized evaluation scenarios.

Our method, RetSTA-7B, outperforms all compared models, including LLMs with significantly larger parameter sizes than 7B. It demonstrates that the proposed paradigm achieves a high-performance standardization model at a relatively low cost, while further validating that RetSTA-7B not only fully understands the correspondence between original reports and standard terminology, but also accurately masters complex expressions encountered in real clinical scenarios.

\textbf{For Chinese reports}, the results are shown in Table~\ref{tab2}. RetSTA-7B significantly outperforms comparison models of similar size as well as those with substantially larger parameter sizes than 7B, demonstrating its strong domain adaptation capability. Notably, RetSTA-7B-Zero also demonstrates impressive performance, indicating that even simple data augmentation, which simulates diverse scenarios in clinical diagnostic reports, can achieve high performance in the task of standardization. This suggests that the proposed data augmentation strategy captures the variability and complexity of real clinical scenarios effectively, providing a cost-efficient approach to model training. Furthermore, the performance gap between RetSTA-7B and RetSTA-7B-Zero highlights the importance of incorporating large-scale bilingual standardized data for further enhancing model capabilities.

\begin{table}[]
\centering
\renewcommand{\arraystretch}{1.2}
\caption{The impact of data augmentation on RetSTA-7B-Zero's standardization performance.}
\label{tab3}
\begin{tabular}{cccccc}
\hline
\textbf{} & \textbf{ Samples } & \textbf{ BLEU-1 } & \textbf{ BLEU-4 } & \textbf{ METEOR } & \textbf{ROUGE-L} \\ \hline
w/ Augmentation                   & 2000                       & 87.11           & 78.84           & 87.56           & 87.82            \\
w/o Augmentation                  & 2000                       & 87.49           & 79.51           & 88.08           & 88.16            \\
w/ Augmentation                   & 10000                      & 91.93           & 87.89           & 92.74           & 92.77            \\ \hline          
\end{tabular}
\end{table}

\begin{table}[]
\centering
\renewcommand{\arraystretch}{1.2}
\caption{Comparison of test results from fine-tuning using Chinese reports only versus bilingual reports.}
\label{tab4}
\begin{tabular}{ccccc}
\hline
\multicolumn{1}{c}{\textbf{}} & \textbf{ BLEU-1 } & \textbf{ BLEU-4 } & \textbf{ METEOR } & \textbf{ ROUGE-L } \\ \hline
RetSTA-7B-Chinese             & 93.43           & 89.84           & 94.24           & 94.29            \\
RetSTA-7B                     & 93.35           & 89.83           & 94.19           & 94.30            \\ \hline
\end{tabular}
\end{table}

\begin{table}[]
\centering
\renewcommand{\arraystretch}{1.2}
\caption{Comparison of test results from fine-tuning using English reports only versus bilingual reports.}
\label{tab5}
\begin{tabular}{ccccc}
\hline
\multicolumn{1}{c}{\textbf{}} & \textbf{ BLEU-1 } & \textbf{ BLEU-4 } & \textbf{ METEOR } & \textbf{ ROUGE-L } \\ \hline
RetSTA-7B-English             & 92.55           & 89.56           & 86.14           & 90.12            \\
RetSTA-7B                     & 92.69           & 89.66           & 86.04           & 90.08            \\ \hline
\end{tabular}
\end{table}

\subsection{Ablation Study}
In Table~\ref{tab3}, we investigate the impact of data augmentation strategies on the standardization performance of the fine-tuned Qwen2.5-7B-Instruct model. We present results under three experimental conditions: fine-tuning using 400 report pairs expanded to 2000 through data augmentation, fine-tuning directly using 2000 report pairs, and the results of RetSTA-7B-Zero. The experiments further validate that simulating complex scenarios in real clinical diagnostic environments to achieve the same data volume can achieve performance comparable to fine-tuning with real data, while significantly reducing costs. This finding provides strong support for efficient and cost-effective model fine-tuning.
In Table~\ref{tab4} and Table~\ref{tab5}, we observe that incorporating English translations into the fine-tuning data does not compromise the model's standardization performance for Chinese diagnostic reports, and vice versa. This finding indicates that the model exhibits strong independence and compatibility when handling bilingual data.

\section{Conclusion}
In this study, we construct a bilingual fundus standard terminology that integrates clinical standard terms and commonly used descriptions, and propose two LLMs for standardization: RetSTA-7B-Zero and RetSTA-7B. RetSTA-7B-Zero effectively handles complex expressions in clinical diagnostic processes, demonstrating exceptional performance in standardization tasks. RetSTA-7B is more powerful, exhibiting superior domain adaptability, significantly outperforming other LLMs in standardization task. For future work, we plan to leverage RetSTA-7B to support downstream tasks, including training text-image foundation models and multimodal LLMs, as well as extending its application to domains beyond ophthalmology.

\begin{comment}  %% removed for anonymized MICCAI 2025 submission.
    
    % The following acknowledgement and disclaimer sections should be removed for the double-blind review process.  
    % If and when your paper is accepted, reinsert the acknowledgement and the disclaimer clause in your final camera-ready version.

\begin{credits}
\subsubsection{\ackname} A bold run-in heading in small font size at the end of the paper is
used for general acknowledgments, for example: This study was funded
by X (grant number Y).

\subsubsection{\discintname}
It is now necessary to declare any competing interests or to specifically
state that the authors have no competing interests. Please place the
statement with a bold run-in heading in small font size beneath the
(optional) acknowledgments\footnote{If EquinOCS, our proceedings submission
system, is used, then the disclaimer can be provided directly in the system.},
for example: The authors have no competing interests to declare that are
relevant to the content of this article. Or: Author A has received research
grants from Company W. Author B has received a speaker honorarium from
Company X and owns stock in Company Y. Author C is a member of committee Z.
\end{credits}

\end{comment}
\bibliography{myReference}
\bibliographystyle{splncs04}

\end{document}